\documentclass[letterpaper, 10 pt, conference]{ieeeconf}  
\IEEEoverridecommandlockouts                              
\usepackage{amsmath}
\usepackage{amsfonts}
\usepackage{amssymb}
\usepackage{dsfont}
\usepackage{graphicx}
\usepackage{wrapfig}
\usepackage{subcaption}
\usepackage{hyperref}
\usepackage{nicefrac}       
\usepackage[ruled]{algorithm2e}
\usepackage{paralist}
\newcommand{\gfm}{\ensuremath{\boldsymbol{\mathcal{F}}_{\phi}}}
\DeclareMathOperator{\argmax}{argmax}
\DeclareMathOperator{\argmin}{argmin}

\title{\textbf{Modelling Generalized Forces with Reinforcement Learning \\ for Sim-to-Real Transfer} }

\author{Rae Jeong, Jackie Kay, Francesco Romano, Thomas Lampe, Tom Rothorl, \\ Abbas Abdolmaleki, Tom Erez, Yuval Tassa, Francesco Nori
\thanks{Authors are with DeepMind London, UK. \{raejeong, kayj, fraromano, thomaslampe, tcr, aabdolmaleki, etom, tassa, fnori\}@google.com. Qualitative results can be found in our supplementary video: \url{https://youtu.be/2diszIMOn6A}}
}

\begin{document}
\maketitle
\thispagestyle{empty}
\pagestyle{empty}

\begin{abstract}
Learning robotic control policies in the real world gives rise to challenges in data efficiency, safety, and controlling the initial condition of the system.
 On the other hand, simulations are a useful alternative as they provide an abundant source of data without the restrictions of the real world.
 Unfortunately, simulations often fail to accurately model complex real-world phenomena.
 Traditional system identification techniques are limited in expressiveness by the analytical model parameters, and usually are not sufficient to capture such phenomena.
 In this paper we propose a general framework for improving the analytical model by optimizing state dependent generalized forces.
 State dependent generalized forces are expressive enough to model constraints in the equations of motion, while maintaining a clear physical meaning and intuition.
 We use reinforcement learning to efficiently optimize the mapping from states to generalized forces over a discounted infinite horizon.
 We show that using only minutes of real world data improves the sim-to-real control policy transfer. We demonstrate the feasibility of our approach by validating it on a nonprehensile manipulation task on the Sawyer robot.
\end{abstract}

\section{Introduction}

Reinforcement learning (RL) \cite{sutton2018reinforcement} algorithms have demonstrated potential in many simulated robotics domains such as locomotion \cite{DBLP:journals/corr/HeessTSLMWTEWER17}, manipulation \cite{Rajeswaran-RSS-18} and navigation \cite{DBLP:journals/corr/abs-1802-01561}.
Yet these algorithms still require large amounts of data, making learning from scratch on real robots challenging \cite{e2e_levine, handeye_levine, riedmiller2018learning}.
Recently, there have been many successes using simulation to speed up robot learning \cite{DBLP:journals/corr/abs-1710-06537, DBLP:journals/corr/SadeghiL16,DBLP:journals/corr/abs-1804-10332, DBLP:journals/corr/abs-1802-09564}.
Recent works have demonstrated that with enough domain randomization~\cite{DBLP:journals/corr/abs-1710-06537}, zero-shot transfer of sim-to-real policies are feasible, even in contact-rich, hybrid discrete-continuous dynamical systems~\cite{DBLP:journals/corr/abs-1710-06537, DBLP:journals/corr/abs-1804-10332}.

A successful sim-to-real transfer method balances the expensive real-world data and inexpensive simulated data to yield sample-efficient performance in the real world.
Dynamics randomization~\cite{DBLP:journals/corr/abs-1710-06537} can yield successful policy transfer, but relies on manual selection of simulation parameters to randomize and does not incorporate real-world data in the learning process.
Model-based RL methods learn the dynamics model from data and tend to be more sample efficient than model-free RL methods, but learning a model still requires high number of real world samples \cite{Clavera2018ModelBasedRL, Nagabandi2018NeuralND}, and may be prohibitive for complex dynamics.
Adversarial domain adaptation methods have the potential to improve the simulation, but have only been demonstrated for adaptation of observations, rather than direct adaptation the system dynamics \cite{DBLP:journals/corr/abs-1709-07857}.
Classical system identification methods provide a sample efficient way to improve the model, but are limited in expressiveness by the model parameters exposed from the analytical models or simulators \cite{DBLP:journals/corr/abs-1803-10371}.

\begin{figure}[t]
\begin{center}
\includegraphics[width=.9\columnwidth]{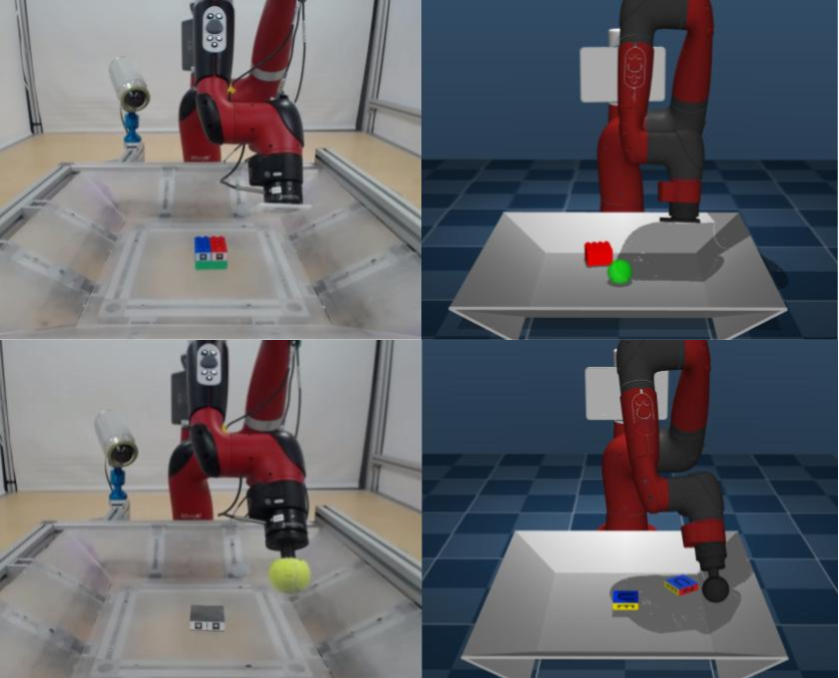}
\end{center}
\caption{Manipulation task setup, on the real robot and in simulation. Top row: 3D position matching task (green dot is the target position). Bottom row: 6D pose matching task (translucent target is the target 6D pose).}
\label{fig:task_setup}
\end{figure}

Nonprehensile robotic manipulation has been studied previously in the form of planar pushing with dynamics randomization and zero-shot sim-to-real transfer \cite{DBLP:journals/corr/abs-1710-06537}.
Policies trained with dynamics randomization has been shown to produce robust policies which can successfully perform real-world nonprehensile robotic manipulation tasks. However, zero-shot sim-to-real transfer using dynamics randomization does not use any real world data to improve the simulator and requires manual tuning of the choices and ranges of the physical parameters to vary in simulation.
The chosen parameters are often very conservative, resulting in a policy that trades robustness to large variation of the model parameters at the expense of overall performance.

Alternatively, the simulated model can be improved to resemble the real world more closely.
System identification techniques have been applied to nonprehensile manipulation through joint system identification and state estimation procedure that accurately models the contact dynamics \cite{DBLP:journals/corr/abs-1803-10371}.
However, system identification methods require the true system dynamics to be in the realm of expressiveness of the analytical model. To get around this, others modelled the complex dynamics of planar pushing by combining both the learned and analytical models \cite{ajayIROS2018}.

Modelling complex dynamics is important for nonprehensile manipulation but is also critical for locomotion. Recent work performed sim-to-real transfer of locomotion tasks by improving the simulation fidelity through system identification, modelling the actuators and latency of the system, which was then used in conjunction with dynamics randomization to train a robust policy for better sim-to-real transfer \cite{DBLP:journals/corr/abs-1804-10332}.
The idea of using both the learned and the analytical model has also been applied to locomotion where the complex actuator dynamics were learned for a quadruped robot \cite{anymal}.

Ground Action Transformation (GAT) \cite{Hanna2017GroundedAT} is a more general method which modifies the actions from the policy trained in simulation.
The modified actions are chosen such that, when applied in simulation, the resulting next state matches the next state observed on the real robot.
Our proposed approach can be seen as a generalization of the Ground Action Transformation method.
We introduce additive generalized forces to the environment instead of modifying the agent's actions.
The optimization of GAT and our method is also different, in that GAT learns an inverse dynamics model in both simulation and the real robot to compute the action transformation, while our method optimizes the objective directly over a discounted infinite horizon through the use of reinforcement learning.

This paper focuses on sim-to-real transfer of robotic manipulation by efficiently utilizing a small amount of real world data to improve the simulation.
Our contribution is twofold.
First, we provide a novel sim-to-real method which learns a state dependent generalized force model using neural networks as expressive function approximators to model the constraint forces of a dynamical system.
Second, we efficiently optimize the state-dependent generalized force model over a discounted infinite horizon through the use of RL. We validate the proposed approach on a nonprehensile robotic manipulation task on the Saywer robot.

This paper is organized as follows. Section \ref{sect:background} introduces the mathematical background on mechanical systems and reinforcement learning problems. Section \ref{sect:contribution} describes how modelling the generalized forces can be used to improve sim-to-real transfer and how these forces can be learned efficiently using reinforcement learning. Finally, Section \ref{sect:experiments} discusses the experimental validation of the proposed approach. Conclusions and perspectives conclude the paper.

\section{Background}
\label{sect:background}

\subsection{Mechanical Systems and Equations of Motion}

In classical mechanics, any rigid body system subject to external forces can be described with the following equations of motion \cite[Ch. 13.5]{Marsden2010}:
\begin{equation}
    \boldsymbol{M}(\boldsymbol{q})\boldsymbol{\dot{v}}+\boldsymbol{c}(\boldsymbol{q}, \boldsymbol{
    \nu})=\begin{bmatrix}\boldsymbol{0}_m\\\boldsymbol{\tau}\end{bmatrix}+ \sum^K_k \boldsymbol{J}_k^\top(\boldsymbol{q})\boldsymbol{f_{k}},
\label{eq:ridgid}
\end{equation}
where $\boldsymbol{q}$ and $\boldsymbol{\nu}$ are the system configuration and velocity respectively. $\boldsymbol{M}$ is the mass matrix, $\boldsymbol{c}$ represents the bias forces, including Coriolis, centrifugal, joint friction and gravitational forces.
$\boldsymbol{\tau}$ denotes the internal, actuation, torques, and $\boldsymbol{0}_m$ is a vector of size $m$ of zeros denoting the unactuated variables.  External 6D forces $\boldsymbol{f_{k}}$ are mapped to generalized forces by the corresponding contact Jacobian $\boldsymbol{J_k}$.

Note that Eq. \eqref{eq:ridgid} is quite general. Depending on the definition of the state variables $\boldsymbol{q}$ and $\boldsymbol{\nu}$, it can be used to represent an articulated system, e.g. a robot arm, a generic free floating object, or both systems.
If the systems are interacting with each other or with the environment, Eq. \eqref{eq:ridgid} needs to be complemented with additional equations describing the contact constraints.
Without making any assumption on the kind of contacts, we can express these constraints as a generic function of the state of the involved systems:
\begin{equation}
\label{eq:contact_constraints}
    h_c(\boldsymbol{q}, \boldsymbol{\nu}) = 0.
\end{equation}
Note that, in the case of rigid contacts, \eqref{eq:contact_constraints} is greatly simplified, but in the following we will consider its most general form.

\begin{figure*}
\begin{center}
\includegraphics[width=2\columnwidth]{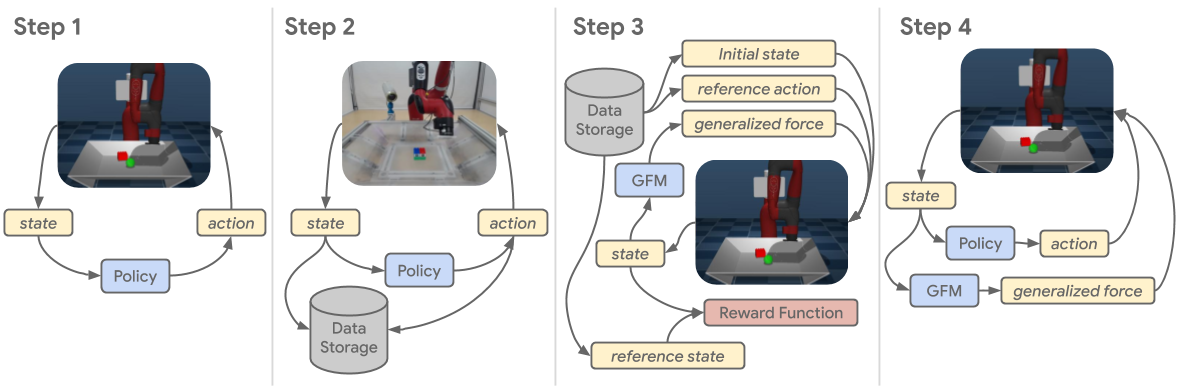}
\end{center}
\caption{Description of the proposed method steps. 1) Train an agent in simulation with the original model parameters. 2) Use the agent trained in step 1 to collect real world data. 3) Learn the generalized force model (GFM): initialize the simulation to the initial state of a real world trajectory and simulate forward, choosing the same actions taken on the real robot. The GFM injects generalized forces into the simulation with the objective of minimizing the difference between the real world and simulation states. 4) Retrain the agent for solving the task of interest using the updated hybrid model.
}
\label{fig:gfm_training}
\end{figure*}

\subsection{Reinforcement Learning}
We consider the reinforcement learning setting with a discounted infinite horizon Markov Decision Process (MDP). An MDP consists of the tuple $(\mathcal{S}, \mathcal{A}, r, P, \mu_0)$: the set of valid states, set of valid actions, the reward function, transition dynamics and the initial state distribution. Let $r(s_t, a_t)$ denote the reward obtained in state $s_t$ of an MDP when executing action $a_t$. The reinforcement learning objective is to find the optimal policy such that:
\begin{equation}
\label{eq:rl}
    \pi^* = \argmax_{\pi \in \Pi} \mathbb{E}_{\pi, P} \Big[ \sum_{t=0}^\infty \gamma^t r(s_t, a_t) | s_0 = \bar{s} \Big],
\end{equation}
where $\gamma$ is the discount factor and $\Pi$ is the set of all possible policies and the expectation with respect to $\pi$ is $\forall t: a_t \sim \pi(\cdot | s_t)$, and the states follow the system dynamics of the MDP, i.e. we have $s_{t+1} \sim P(\cdot | s_t, a_t)$ where $P(s_{t+1} | s_t, a_t)$ denotes the transition probability.

With this notation in place, we can define the action-value function, which evaluates the expected sum of rewards for a policy $\pi$ with a one-step Bellman equation of $Q^\pi(s,a)$:
\begin{align}
\label{eq:q}
    Q^\pi(s_t, a_t) &= r(s_t, a_t) + \gamma \mathbb{E}_{\pi, P} \Big[Q^\pi(s_{t+1}, a_{t+1}) \Big].
\end{align}

The action-value function is learned for off-policy reinforcement learning during the \textit{policy evaluation} step to evaluate the policy. The policy itself is improved using the result of policy evaluation in the \textit{policy improvement} step. Iteratively applying these two steps is referred to as \textit{policy iteration} \cite{sutton2018reinforcement}.

\section{Modelling Generalized Forces with RL}
\label{sect:contribution}

Our method uses a hybrid approach: the analytical model of a mechanical system is supplemented with a parametric function representing additional generalized forces that are learned using real world data.
Consider the dynamics equation in Eq. \eqref{eq:ridgid}. In addition to the actuation and external forces, we add a state-dependent generalized force term, i.e.
\begin{equation}
    \boldsymbol{M}\boldsymbol{\dot{v}}+\boldsymbol{c}=\begin{bmatrix} \boldsymbol{0}_m \\ \boldsymbol{\tau} \end{bmatrix} + \sum^K_k \boldsymbol{J}_k^\top\boldsymbol{f_{k}} + \boldsymbol{\mathcal{F}}_{\phi}(\boldsymbol{q}, \boldsymbol{\nu}).
\label{eq:eom_gfm}
\end{equation}

We represent the contact-aware system dynamics in Eqs. \eqref{eq:eom_gfm} and \eqref{eq:contact_constraints} as a hybrid discrete dynamics model ${s_{t+ 1}} = f_{\boldsymbol{\theta}}({s_t}, {a_t})$, parameterized by vector $\boldsymbol{\theta}$, which takes as input the current state and action and outputs the prediction of the next state ${s}_{t+1}$.
We want to find the parameter vector $\boldsymbol{\theta}^*$ that minimizes the dynamics gap along a horizon $T$, that is:
\begin{equation}
    \boldsymbol{\theta}^* = \argmin \sum_{t=0}^{T} ||{x}_{t+1} - f_{\boldsymbol{\theta}}({s}_t, {a}_t)||^2
\label{eq:sysid}
\end{equation}
where ${x}_{t+1}$ represents the resulting next state in the real system.
In contrast to other hybrid modelling approaches that learn the residual state difference, we choose to model the gap with generalized forces.
This choice has two benefits:
\begin{inparaenum}[i)]
\item generalized forces are applied directly to the dynamics equation, that is, the next state resulting from the application of Eqs. \eqref{eq:eom_gfm}, \eqref{eq:contact_constraints} is a physically consistent state;
\item it is easier to impose constraints on the learned generalized forces as they have a physical correspondence.
\end{inparaenum}

It is worth noting that the resulting hybrid model is non-differentiable, as Eq. \eqref{eq:contact_constraints} is in general non-differentiable.
By consequence, the function $f_{\boldsymbol{\theta}}$ is not differentiable with respect to its arguments and the optimal generalized forces $\boldsymbol{\mathcal{F}}^*_{\phi}$ are not trivial to obtain using classical methods.
However, the formulation of optimizing a state-dependent model interacting with a non-differentiable component over a horizon strongly resembles the reinforcement learning objective in Equation \ref{eq:rl}.
We therefore reformulate the original problem of optimizing the objective in Equation \ref{eq:sysid} as a reinforcement learning problem.

Define ${z}_i = \{{x_0, a_0, x_1, a_1, x_2}, \cdots\}$ a trajectory consisting of a sequence of states and actions collected on the real system, and $\boldsymbol{X} = \{{z}_i\}_{i=0}^{N}$ the corresponding dataset of all the acquired real world trajectories.
Let us define the following reward function:
\begin{equation}
   r_f({x_t}, {s_t}) = e^{-||\Delta({x}_t, {s}_t)||^2},
\label{eq:expr}
\end{equation}
with $\Delta({x}_t, {s}_t)$ a suitable distance function for measuring the distance between ${x}_t$ and ${s}_t$.
At the beginning of each training episode, we randomly sample an initial state ${x}_0 \in \boldsymbol{X}$, and we initialize the internal dynamics model to ${x}_0$.
The sum of the rewards in Equation \ref{eq:expr} is maximized with the RL objective in Equation \ref{eq:rl}, resulting in the following objective:
\begin{equation}
\label{eq:gfmrl}
    \boldsymbol{\mathcal{F}}_{\phi}^* = \argmax_{\boldsymbol{\phi}} \mathbb{E}_{\boldsymbol{\mathcal{F}}_{\phi}, P_S, \boldsymbol{X}} \Big[ \sum_{t=0}^\infty \gamma^t r_f({x}_t, {s}_t) | {s}_0 = {x}_0 \Big],
\end{equation}
where $P_S$ is the transition probability distribution of the unmodified dynamic model given in Eqs. \eqref{eq:ridgid},\eqref{eq:contact_constraints}.

The resulting generalized force model (GFM), $\boldsymbol{\mathcal{F}}_{\phi}$, is then used as the transition model $P_f$ for the hybrid model \eqref{eq:eom_gfm} and \eqref{eq:contact_constraints}.
We then train the policy for a control task of interest using the hybrid model with the transition probability distribution $P_f$, resulting in the following reinforcement learning objective:
\begin{equation}
\label{eq:finalrl}
    \pi^* = \argmax_{\pi \in \Pi} \mathbb{E}_{\pi, P_f} \Big[ \sum_{t=0}^\infty \gamma^t r({s}_t, {a}_t) | {s}_0 = \bar{s} \Big].
\end{equation}

To optimize both Equation \ref{eq:gfmrl} and \ref{eq:finalrl}, we chose the \textit{policy iteration} algorithm called Maximum a Posteriori Policy Optimization (MPO)~\cite{mpo}, which uses an expectation maximization style \textit{policy improvement} with an approximate off-policy \textit{policy evaluation} algorithm Retrace~\cite{retrace} which estimates the action-value function given in Equation \ref{eq:q}.

Fig. \ref{fig:gfm_training} shows the proposed method, highlighting the different steps from the first policy training in the unmodified simulator ($P_S$, step 1) to the final policy trained on the updated simulator model ($P_f$, step 4).

\begin{figure}[t]
\begin{center}
\includegraphics[width=\columnwidth]{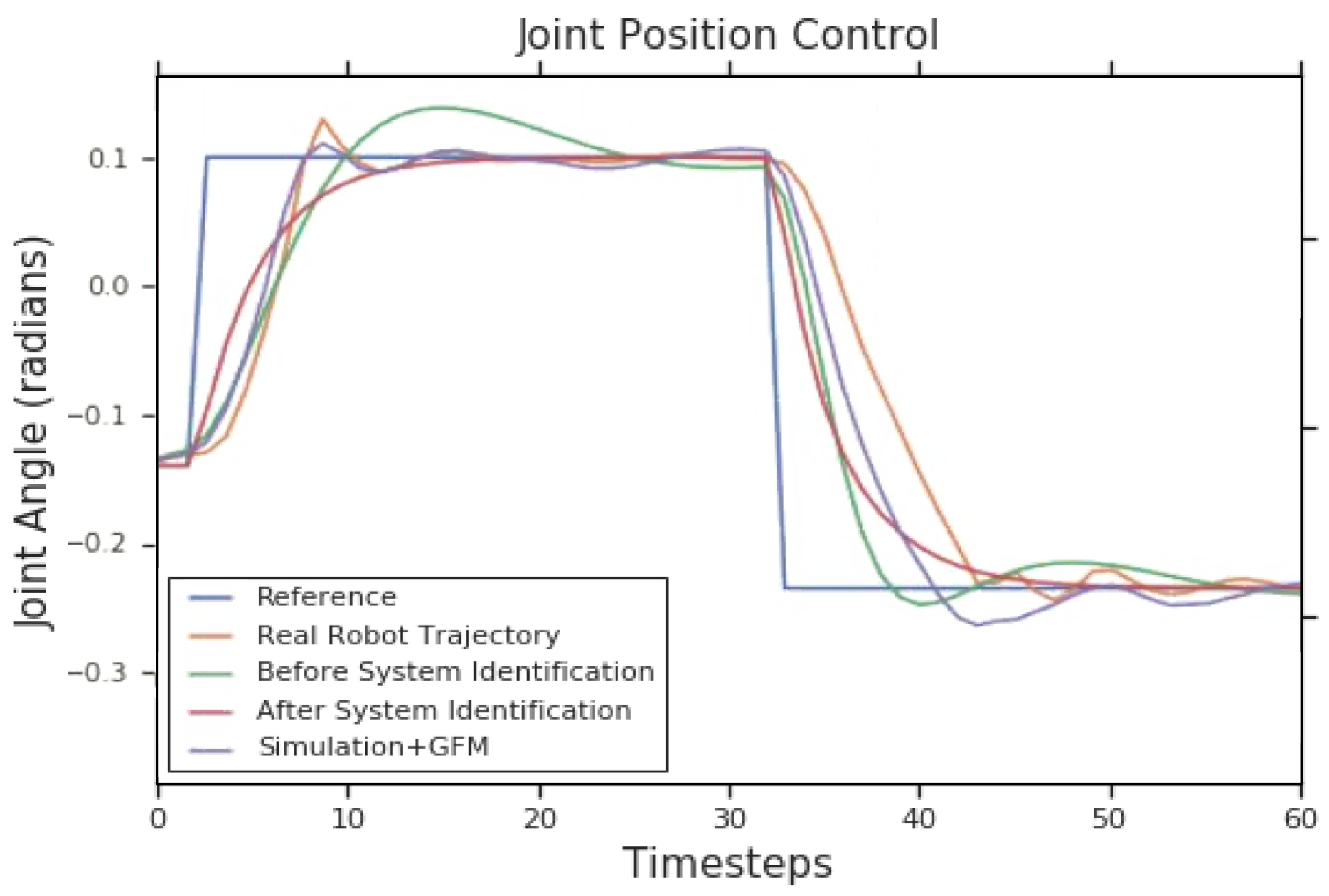}
\end{center}
\caption{Joint position control experiment. Pictured are the position setpoint reference trajectory for the base joint of the arm, the position of the real robot after following the command, the prediction from the model with default parameters, after system identification and from the hybrid model.}
\label{fig:joint_position_control}
\end{figure}

\section{Experimental Results and Discussion}
\label{sect:experiments}
In this section we present the results of the experiments we performed to validate the proposed sim-to-real transfer method.
The experiments have been devised to answer to the following questions.
\begin{inparaenum}[i)]
\item Does the use of the GFM improve the modelling capacity with respect to an analytical model with parameters obtained from system identification?
\item How does the training of GFMs scale with an increase in real world data?
\item What is the impact of GFMs on policy training in simulation?
\item Does the use of GFMs improve sim-to-real transfer for nonprehensile robotic manipulation tasks?
\end{inparaenum}

Referring to Eq. \eqref{eq:eom_gfm}, the configuration of our experimental system is $\boldsymbol{q} \in \mathbb{R}^7 \times \mathbb{R}^3 \times SO(3)$ with the velocity $\boldsymbol{\nu} \in \mathbb{R}^{7 + 6}$.
The actuated torques are $\boldsymbol{\tau} \in \mathbb{R}^7$ with the learned generalized forces $\gfm$ $\in \mathbb{R}^{7+6}$.

\begin{figure}[t]
\begin{center}
\includegraphics[width=.9\columnwidth]{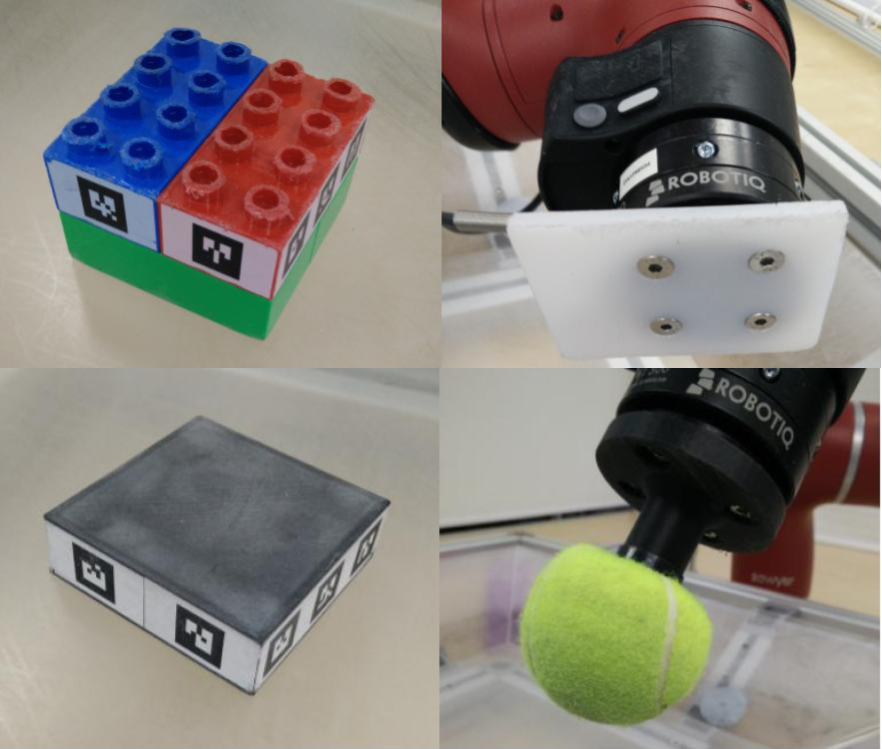}
\end{center}
\caption{Top row: object and plate end effector for the 3D position matching task. Bottom row: object and ball end effector for the 6D pose matching task. Objects 6D poses are tracked using the AR tags on the sides of the object.}
\label{fig:object_n_ee}
\end{figure}

The position and velocity of the robot joints are obtained by using the local robot sensors, i.e. the joint encoders.
The object is tracked by using a fiduciary marker system based on AR tags, see Fig. \ref{fig:object_n_ee}. We track the object independently with three cameras and then merge the information into a single 6D estimation value.
To obtain the velocity, we perform a finite differentiation of the estimated position.
The robot is controlled at $20\mathrm{Hz}$ while the simulation is integrated every $2\mathrm{ms}$.

With reference to the algorithm outlined in Sect. \ref{sect:contribution}, we define the distance function $\Delta({x}_t, {s}_t)$ as follows:
\begin{equation}
\Delta({x}^{(i)}_t, {s}^{(i)}_t) =
\begin{cases}
    {x}^{(i)}_t - {s}^{(i)}_t \quad \text{if }{x}^{(i)}_t \in \mathbb{R}^l \\
    {s}^{(i)}_t \star {x}^{{(i)}*}_t  \quad \text{if }{x}^{(i)}_t \in SO(3),
  \end{cases}
\end{equation}

\subsection{Experimental Setup}
\label{sect:exp_setup}
Fig. \ref{fig:task_setup} shows the real and simulated setup used in our experiments.
The platform is a 7 degrees of freedom (DoF) Rethink Robotics Sawyer robotic arm.
A basket, where the objects lie and move, is placed in front of the robot.
Depending on the specific experiment, the object and the end effector of the robot can change.
The full hardware setup, i.e. robot, basket and objects, are modelled in simulation using the MuJoCo simulator \cite{todorov2012mujoco} where we represented the orientation component by using unitary quaternions, $\star$ is the quaternion product and ${x}^{{(i)}*}_t$ denotes the complex conjugate of the quaternion.
We define the following limits for the learned generalized forces:
\begin{itemize}
    \item $5\mathrm{Nm}$ for the forces acting on the robot joints,
    \item $0.03 \mathrm{N}$ for the forces acting on the translational part of the object and
    \item $0.03 \mathrm{Nm}$ for the forces acting on the rotational part of the object.
\end{itemize}
While learning the GFMs, we also provide as observation to the RL algorithm the 5 previous states of the dynamics simulation.

Both the GFM and the control policy are feed-forward neural networks with two layers, each with a size of 200. The neural network outputs a Gaussian distribution with the mean $\mu$ and diagonal Cholesky factors $A$, such that $\Sigma = AA^T$. The diagonal factor $A$ has positive diagonal elements enforced by the softplus transform $A_{ii} \leftarrow \log(1 + \exp(A_{ii}))$ to ensure positive definiteness of the diagonal covariance matrix.

\subsection{Joint Position Control}
To examine the GFM's modelling capabilities, we collect trajectories of the robot controlled in position mode, following a pre-defined waypoint-based trajectory.
No interactions with the object or the basket took place in this experiment.
MuJoCo by default exposes two parameters for modelling the actuator, namely the actuator gain and damping.
We identify these parameters by using SciPy \cite{scipy} Sequential Least Squares Programming.
We then train our proposed hybrid model using the approach described Sect. \ref{sect:contribution}.

We report the comparison of the trajectories resulting from the different models in Fig. \ref{fig:joint_position_control}.
We can notice that both the GFM and the parameters-identified model improve upon the default simulation model.
However, the GFM hybrid model is able to capture the complexity of the real data better than the simulator's simplified actuator model alone. 
This also demonstrates the generality of our GFM hybrid model approach, as no specific knowledge about actuator dynamics are needed for training the GFM.
However, it is worth noting that the GFM hybrid model is a local approximation around the real data used for training.

\begin{figure}[t]
\begin{center}
\includegraphics[width=3.1in]{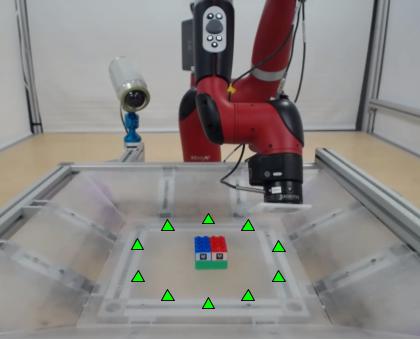}
\end{center}
\caption{The 10 target positions for the position matching task are shown as the green triangles.}
\label{fig:target_locations}
\end{figure}

\subsection{Position Matching Task}
\label{sect:position_matching_task_description}
To validate and assess quality of the proposed method, we use a position matching task for the free-moving object.
The goal of this task is to move the object to a specific target location.
The robotic arm is equipped with an acrylic plate as its end effector as can be seen in Fig. \ref{fig:object_n_ee}.

We provide the state of the robot, the pose of the object, and the target position of the brick as observations to the agent, as described in Sect. \ref{sect:exp_setup}.
We provide a history of observations by providing to the agent the last $10$, including the current, observations.
Each episode is composed of a maximum of $400$ transitions (corresponding to $20\mathrm{s}$).
The episode is terminated, and considered a failure, if the end-effector of the robot leaves its workspace.

The agent's actions consist of joint velocities, representing reference setpoints for the robot's low-level joint velocity controller.
We decided to limit the maximum joints velocity to 5 percent of the allowed range of velocities.
Two reasons motivated this choice:
\begin{inparaenum}[i)]
\item lower velocity implies safer action from the robot while interacting with the object and the environment, and,
\item lower velocity for the object whilst under interaction with the robot allows better pose estimation from the AR tags system.
\end{inparaenum}

We evaluate the agent's performance on the task with a binary success-failure criterion for 5 trials. 
An episode is a success if the error between the object's position and the target is less than $2.5\mathrm{cm}$ for $10$ consecutive time steps, i.e. $0.5\mathrm{s}$.
Each trial consists of 10 target positions, making the total number of target positions attempted for each policy to $50$.
The initial pose of the object is reset to the center of the basket at the beginning of every attempt.
The target locations are fixed and are radially positioned around the center of the basket -- See Figure \ref{fig:target_locations}.

\begin{figure}[t]
\begin{center}
\includegraphics[width=\columnwidth]{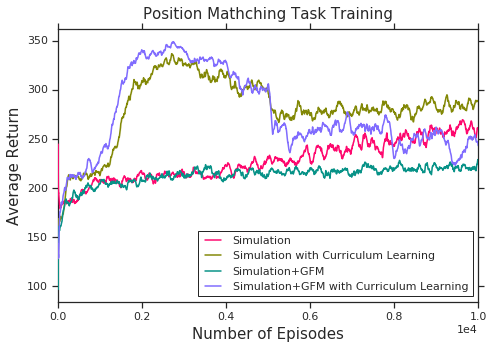}
\end{center}
\caption{Average return when learning the position matching task with and without the use of curriculum and GFM. The use of curriculum results in a non-monotonic learning curves, because throughout the course of training the control authority of the policy is reduced.}
\label{fig:position_matching_training}
\end{figure}

\subsection{Ensemble of Hybrid Models}
We first examine how the training of GFMs scales with the increase in real world data.

We train a policy for the position matching task using the default simulator model as shown in step 1 of Fig. \ref{fig:gfm_training}, and we use this policy to collect a dataset of $10$ trajectories on the real robot.

We then train two GFMs using different amounts of real world trajectories.
The first one uses 5 trajectories, while the second one uses all 10 trajectories.
The second GFM faces a more challenging modelling problem as it has to learn to fit more data points from the real world.
In our GFM training, using 5 trajectories on average resulted in around 10\% higher average return when compared to the GFM that used 10 trajectories.

In practice, given that we would like to use more than 5 real world trajectories to improve the dynamic model, we decide to employ an ensemble of GFMs instead of increasing the number of trajectories used during the GFM learning.
All the following experiments are trained with the fixed number of 5 trajectories per model.

\subsection{Training Policies on Hybrid Models}

The additional complexity introduced by the hybrid model has an impact in the learning performance of the final policy.
Fig. \ref{fig:position_matching_training} shows that the additional $\gfm$ term significantly slows down training when compared to the original unmodified model.
This is expected, as GFM essentially injects force perturbations which makes the task harder to solve.
To reliably obtain the optimal policy when training with the hybrid model, we employ a simple curriculum learning strategy.
At the beginning of the training, we allow higher control authority to facilitate exploration and task completion, while progressively lowering it towards the end.
We implemented the curriculum by introducing an additional multiplicative gain to the controlled joint velocities.
We started with a gain of $6$, progressively reducing it to $1$ by unitary steps every $1000$ episodes.
Fig. \ref{fig:position_matching_training} shows the training curves for policy learning on both the unmodified dynamic model and the hybrid model when using the curriculum strategy.
We can clearly see that the curriculum improves the speed and the final performance of the policy.

\begin{table}[t]
\caption{Position Matching Task Evaluation}
\begin{center}
\begin{tabular}{|l|c|c|}
\hline
\textbf{Models} & \textbf{Task Success} & \textbf{Std. Dev.}\\
\hline
\hline
Original model & 38\% & 3.4\\
\hline
Original model with Curriculum & 62\% & 3.4\\
\hline
Random Force Perturbations & 70\% & 3.2\\
\hline
\hline
Hybrid Model (1 model in ensemble) & 58\% & 3.5\\
\hline
Hybrid Model (3 models in ensemble) & \textbf{84\%} & 2.6\\
\hline
Hybrid Model (5 models in ensemble) & 74\% & 3.1\\
\hline
\end{tabular}
\end{center}
\label{table:results}
\end{table}

\subsection{Position Matching Task: Experimental Results}
This section reports the results of the sim-to-real policy transfer for the nonprehensile position matching task when using both the model ensemble and curriculum learning strategy discussed in the previous sections.

Table \ref{table:results} summarizes the results, reporting the average success percentage and the standard deviation.
We compare our method to 3 baseline methods which are
\begin{inparaenum}[i)]
\item policy trained with the original dynamic model,
\item policy trained with the original model and using curriculum learning and,
\item policy trained with random perturbations of generalized forces and using curriculum learning.
\end{inparaenum}
As expected, the policy trained with the original model performs poorly due to the dynamics gap resulting in 38\% success.
However, the default simulation trained with our curriculum learning performs significantly better with 62\% success.
Given that the curriculum learning induces time-varying actuator gains, this has a similar effect to applying domain randomization.
Lastly, the addition of random generalized force perturbations on top of our curriculum learning results in 70\% success.

The lower part of Table \ref{table:results} reports the performance of transferred policies trained with the hybrid model for a different number of model ensembles (in parentheses).
While the use of one GFM leads to an improvement over the original model, it does not perform better than the other two baselines.
Given that a single GFM only uses 5 real world trajectories, it is plausible that the hybrid model is over-fitting to those particular trajectories.
The use of 3 GFMs significantly improves the sim-to-real transfer and results in 84\% success.
Using 5 GFMs still outperforms all of the baselines but results in a degradation of performance when compared to using only 3 GFMs.
Indeed, while using more GFMs allows us to utilize more real world data, it also makes the task more challenging, even compared to the real world.
This results in a more conservative controller which is not as optimal as when we have the right balance of GFMs to model the real world.

\begin{table}[t]
\caption{Pose Matching Task Evaluation}
\begin{center}
\begin{tabular}{|l|c|c|}
\hline
\textbf{Models} & \textbf{Task Success} & \textbf{Std. Dev.}\\
\hline
\hline
Original model & 18\% & 2.7\\
\hline
Hybrid Model (3 models in ensemble) & \textbf{44\%} & 3.5\\
\hline
\end{tabular}
\end{center}
\label{table:results1}
\end{table}

\subsection{6D Pose Matching Task}
\label{sect:6d_pose}
Finally, we provide the results for sim-to-real policy transfer for a much more challenging nonprehensile manipulation task.
The task is similar to the one described in Sect. \ref{sect:position_matching_task_description}, but this time we consider also the orientation of the object.
The task and learning setup is exactly the same as the one described in Sect. \ref{sect:position_matching_task_description}, except for the following changes. The episode duration is $600$ transitions, i.e. $30\mathrm{s}$. During the evaluation procedure we consider an attempt as successful if the position error is less than $5\mathrm{cm}$ and the orientation error around the vertical axis is less than $20$ degrees.

Table \ref{table:results1} presents the results of the sim-to-real transfer for the 6D pose matching task. As for the previous experiment, our method is able to significantly improve over the baseline of 18\% to 44\% success.
The success rate is lower when compared to the previous experiment, but this applies to both the policy learned on the original model and on the hybrid model, showing the complexity of the task.

\section{Conclusions}

In this work we presented a novel sim-to-real method that aims to close the sim-to-real gap by modelling and learning the state dependent generalized forces that capture this discrepancy.
We validated the proposed approach by performing sim-to-real transfer for nonprehensile robotic manipulation tasks.
We showed that our method improves the performance of sim-to-real transfer for 3D position matching and 6D pose matching task of different objects. While the results reported in Sect. \ref{sect:6d_pose} on 6D pose matching showed that our approach yields a considerable improvement with respect to the default model, there are still rooms for improvements. We thus want to analyze and improve on this task further.

When controlling interaction forces, it is common to choose torque control as the low level controller. Despite this classic choice, the use of torque control is not common in reinforcement learning.
Future work could investigate if the use of our proposed generalized force model helps in learning an interaction task while controlling the robot by using torque control.

\addtolength{\textheight}{0cm}
\bibliographystyle{IEEEtran}
\bibliography{IEEEabrv,reference}

\end{document}